\documentclass[journal]{IEEEtran}
\let\labelindent\relax
\usepackage{graphicx}
\usepackage{subcaption}
\usepackage{eqlist}
\usepackage{amsfonts}
\usepackage{tabularx}
\usepackage{booktabs}
\usepackage{amssymb}
\usepackage{amsmath}
\usepackage{enumitem}
\usepackage{threeparttable}
\usepackage[lined, ruled, linesnumbered, commentsnumbered]{algorithm2e}
\usepackage{lipsum}

\usepackage[usenames,dvipsnames]{xcolor}
\usepackage{flushend}
\usepackage{comment}
\newcommand{\jimmy}[1]{\textcolor{red}{#1}}

\usepackage{multirow}
\usepackage{lipsum}
\newcommand\dunderline[3][-2pt]{{%
  \setbox0=\hbox{#3}
  \ooalign{\copy0\cr\rule[\dimexpr#1-#2\relax]{\wd0}{#2}}}}

\begin{document}
%
\title{Human-in-the-loop Robotic Manipulation Planning for Collaborative Assembly}

\author{Mohamed Raessa, \IEEEmembership{Student Member, IEEE}, Jimmy Chi Yin Chen, \\
Weiwei Wan, \IEEEmembership{Member, IEEE}, and Kensuke Harada,
\IEEEmembership{Member, IEEE}
\thanks{Mohamed Raessa, Weiwei Wan, and Kensuke Harada are with
Graduate School of Engineering Science, Osaka University, Japan.
Jimmy Chen is with the University of California, Santa Cruz, USA.
Correspondent author: Weiwei Wan,
E-mail: wan@sys.es.osaka-u.ac.jp}
}

\markboth{Journal of \LaTeX\ Class Files,~Vol.~6, No.~1, January~2007}%
{Shell \MakeLowercase{\textit{et al.}}: Bare Demo of IEEEtran.cls for Journals}

\maketitle

\begin{abstract}
This paper develops a robotic manipulation planner for human-robot collaborative assembly. Unlike previous methods which study an independent and fully AI-equipped autonomous system, this paper explores the subtask distribution between a robot and a human and studies a human-in-the-loop robotic system for collaborative assembly. The system distributes the subtasks of an assembly to robots and humans by exploiting their advantages and avoiding their disadvantages. The robot in the system will work on pick-and-place tasks and provide workpieces to humans. The human collaborator will work on fine operations like aligning,  fixing,  screwing, etc. A constraint-based incremental manipulation planning method is proposed to generate the motion for the robots. The performance of the proposed system is demonstrated by asking a human and the dual-arm robot to collaboratively assemble a cabinet. The results showed that the proposed system and planner are effective, efficient, and can assist humans in finishing the assembly task comfortably.
\end{abstract}
\def\abstractname{Note to Practitioners}
\begin{abstract}
This paper was motivated by assembling a cabinet. The assembling process involves several pick-and-place, reorientation, regrasp, alignment, peg-in-hole, and screwing sub-tasks. The system distributes these subtasks to robots and humans by exploiting their advantages and avoiding their disadvantages. The robot used is a dual-arm robots with two parallel grippers. Thus, a suction cup tool is used for the pick-and-place of thin objects. Soft-finger contact constraints are considered to assure safe manipulation. Human ergonomics are included as a quality for optimizing the handover between robot arms and robot and human.  The proposed planner and system show satisfying user experience and expedite the cabinet assembly. It is expected to help relax the labor-intensive manufacturing process in the future.
\end{abstract}



%
\IEEEpeerreviewmaketitle

\section{Introduction}
%
%
%
%
\IEEEPARstart{H}{uman} labor is becoming costly in manufacturing. There are high demands for replacing humans with robots. Nevertheless, although robotic planning and control have been developed for several decades, it remains difficult to flexibly deal with fine operations using a fully autonomous and general-purpose robot. Thus, in this work, instead of directly replacing humans with robots, we develop a system that allows human and robot to work together to finish assembly tasks. Humans will work on fine operations like aligning, fixing, screwing, etc. Robots will work on the pick-and-place tasks and provide workpieces to humans while considering human ergonomics and the assembly progress. With the help of manipulation planning considering various constraints, robots can assist humans in finishing the assembly task comfortably and efficiently.
\begin{figure}[!tbp]
    \centering
    \includegraphics[width = .97\linewidth]{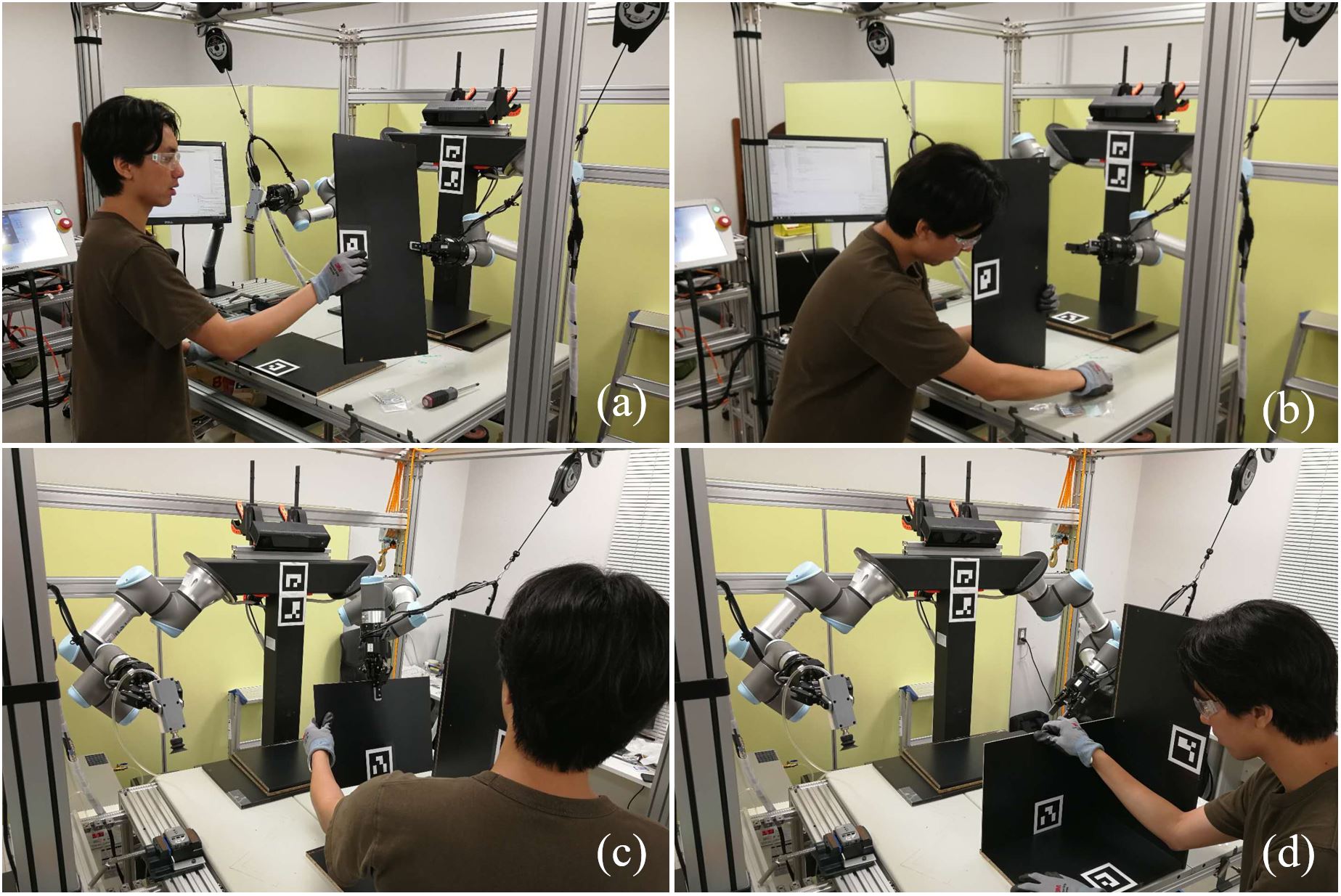}
    \caption{A human assembling a cabinet with the help of a robot. The robot performs the pick-and-place subtasks and provides workpieces to humans. The human works on fine operations like aligning, fixing, screwing, etc. While the human performs the last subtask, the robot system may manipulate and prepare the next workpieces to accelerate the whole assembly process.}
    \label{teaser}
\end{figure}
Previously in industry, the assembly task was either done by an independent industrial robot in a workcell or a human worker. In the case of an independent industrial robot, the system is usually developed by some experienced system integrators with specially designed fixtures and robot end-effectors. There are lots of disadvantages for the previous systems. They are costly, difficult to reconfigure for new objects, and labor-intensive. For these reasons, researchers have been studying AI (Artificial Intelligent) systems that can automatically recognize objects \cite{hashimoto2017}, plan motions \cite{latombe2012robot}, and perform the fine operations \cite{cheng2018sensor}\cite{su2018learning}. Despite the popularity in academia, none of the problems can be solved robustly\jimmy{}, making the implementation of an AI system impractical. In this work, instead of implementing an AI system, we explore the subtask distribution between a robot and a human and study a human-in-the-loop robotic system for collaborative assembly. The system combines certain functions of a traditional industrial robot and a full AI system. The system distributes the subtasks of an assembly to robots and humans by exploiting their advantages and avoiding their disadvantages. The robot in the system will work on pick-and-place tasks and provide workpieces to humans. The human collaborator will work on fine operations like aligning,  fixing,  screwing, etc.

Our main contribution is in the robotic manipulation planning method considering human ergonomics and various constraints. Fig.\ref{teaser} exemplifies the developed system. The robot we used is a dual-arm robot with two Universal Robotics 3 6-DoF robots and Robotiq 85 two-finger grippers. FT sensors are mounted between the robots and the grippers at the robots’ end flanges. The system uses RGB cameras and AR markers to recognize workpieces, uses constrained motion planning to perform manipulation, and provides the workpieces to the human. During the manipulation planning and handover, the robot takes the comfortableness of the human receiving the workpieces into consideration and assists the human in performing fine assembly subtasks. While the human performs a subtask, the robot system may manipulate and prepare the following workpieces to accelerate the whole assembly process. Also, the robotic end-effectors are a pair of general parallel grippers. They are advantageous in the generality but are difficult to control since (i) the pose of a grasped object may easily change depending on the inclination angle, (ii) they cannot pick up thin objects too. To overcome the disadvantages, we extend a conventional robot motion planner by considering soft finger contact constraints.

In the main content of this paper, we present the details of the system implementation and the planning algorithm, as well as demonstrate the performance of the system by asking a human and the dual-arm robot to collaboratively assemble a cabinet. The robot provides cabinet boards to human with proper orientations. The human receives the boards and finishes the assembly task. We compare the performance of the system with two other methods: robotic pick-and-place without considering ergonomics and single-human assembly. The results showed that the developed system with the proposed planning method is effective, efficient, and can assist humans to finish the assembly task comfortably.

The paper is organized as follows. Section II presents the related work. Section III presents an overview of the system and the workflow of the proposed method. Section IV presents the details of determining the planning goals for the constrained robotic manipulation planning. Section V presents the details of the constrained robotic manipulation planning. Section VI presents the implementations of the handover and other miscellaneous stuff. Section VII presents the experiments and analysis. Section VIII draws conclusions and discusses future work.

\section{Related Work}
We review the related work in two aspects: Human-robot collaboration and constrained manipulation planning, with special focus on multi-modal motion planning and the soft-finger contact constraint.

\subsection{Human-robot collaboration}
Most of the human-robot collaboration studies focus on the force control problem. For example, early studies like Kosuge et al. \cite{kosuge1997control} developed model-based methods to control a manipulator that collaboratively handle an object with a human. Harada et al. \cite{harada2007real} used a humanoid robot and real-time gait planning with force feedback from the robot hands to collaboratively move a table with a human. Both of them used force sensors and concentrated on force control. Modern studies tend to use learning methods to learn various control parameters. For example, Rozo et al. \cite{rozo2016learning} developed a system that learned physical behaviors from human demonstration.

More recently, with the development of task and motion planning methods \cite{kambhampati1991combining} \cite{kaelbling2010hierarchical}, researchers began to study human-robot collaboration by automatically generating helpful robotic manipulation. For example, Luo et al. \cite{luo2018unsupervised} developed algorithms to predict human reaching and proposed a planner to plan collaborative robot motion while avoiding collision with humans. Chen et al. \cite{chen2018manipulation} developed a manipulation algorithm for a robot to hold an object for a human while resisting large external force from a human. Maeda et al. \cite{maeda2017probabilistic} developed an algorithm to predict the intention of humans and use that to plan robot trajectory and coordinate the human-robot collaboration accordingly. The focus was on inferring human intentions. Wisley \cite{chan2016research} developed a robot controller that can implement human-like and human-preferred handover, considering several quality measurements like object affordance, human intention, load force, etc.

In this work, we develop a human-robot collaborative assembly system, with special focus on the manipulation planning aspect of the robot. The difference from previous studies is two-fold: On one hand, the goal of the planning is derived by evaluating the comfortableness of handover to a human. On the other hand, our manipulation planning considers both the constraints of a human and the constraints from the manipulated objects. To the best of our knowledge, this work is the first study that combines handover and manipulation planning while considering human in the loop.

\subsection{Constrained robot motion planning considering multi-modality and soft-finger contact}
Generating robotic motion considering various constraints, especially the multi-modality of the planning space and the soft-finger contact constraints, is also related to this work.
\begin{figure*}[!htbp]
    \centering
    \includegraphics[width = \textwidth]{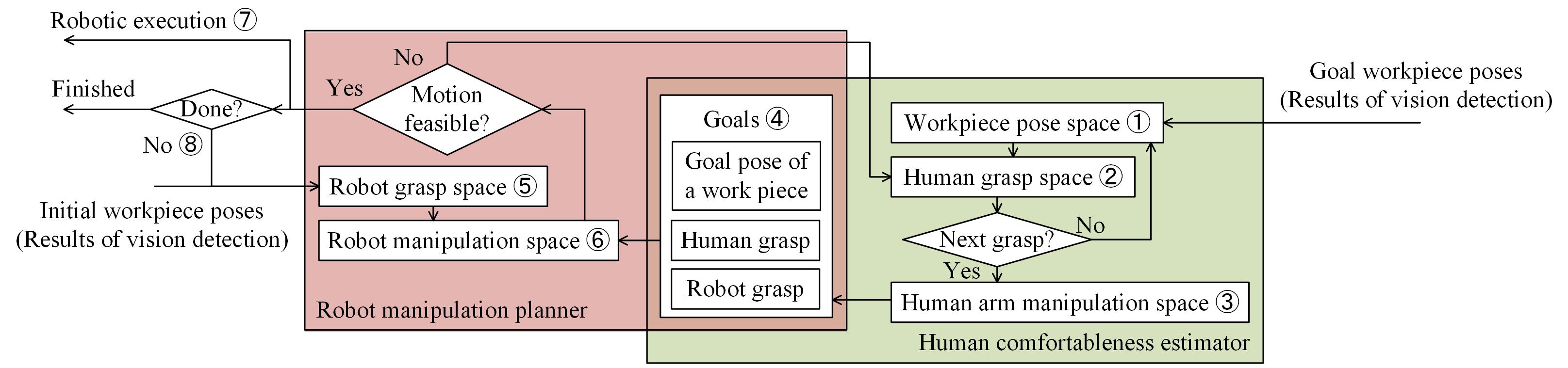}
    \caption{Overview of the proposed human-in-the-loop collaborative assembly system. The flow starts from the right by the target pose of assembly task as an input to the human comfortableness estimator. Several poses for the workpiece are sampled and searched for the availability of human grasp and arm comfortableness. Then, the set of the human comfortable workpiece poses are passed to the robot manipulation planning. The set is checked for available robot grasps and motion feasibility. If feasible motion sequences can be generated for the task, the planning is complete. Otherwise, the flow returns to the workpiece pose space until a candidate workpiece pose that satisfies both the human comfortableness and robot reachability is obtained.}
    \label{workflow}
\end{figure*}

Motion planning under constraints was initially studied as a special case of manipulators redundancy resolution using various optimization methods \cite{siciliano1990kinematic}\cite{martin1989resolution}\cite{chiaverini2008kinematically}. The focus was to realize better utilization of redundant degrees of freedom in the inverse kinematics problem solution. For example, end-effector constraints like the Task-Oriented Constraints \cite{khatib1987unified}\cite{mason1981compliance} or the Task Space Region \cite{Berenson-2011-7265} were used to generate task-related robot motion. They were proved useful in many tasks such as welding \cite{ahmad1989coordinated} and door opening \cite{stilman2007task}. Especially for the human-robot collaboration system studies in this paper, the focus is on reorienting objects by planning across multi-modal configuration spaces. The concept was seminally studied by Koga at al. \cite{koga1994multi} and was conceptualized by Hauser \cite{hauser2010multi}\cite{hauser2011randomized,ng2012multi}. Our group at Osaka University also developed an integrated task and motion planning system considering multi-modal configuration spaces for object reorientation \cite{wan2019regrasp}\cite{raessa2019teaching}. More recently, the technique is extended to heterogeneous planning using both prehensile and nonprehensile grasps \cite{lee2015hierarchical}\cite{fan2019modal}\cite{hangslid19}, or both in-hand manipulation and multi-hand handover \cite{cruciani2018dexterous}.
Another special constraint we are interested in is the soft-finger contact constraint, which affects the stability of grasp and the success of manipulation. The soft finger differs from other types of contacts in its ability to exert friction torque on the object \cite{cutkosky1989grasp}. The friction torque makes the grasp more resistive to the external disturbance. The soft-finger contact was initially described by \cite{johnson1987contact} and was later solidified by \cite{howe1996practical}. Most previous studies used the constraint to find stable grasps for robotic hands \cite{harada2014stability}\cite{ciocarlie2007soft}.

In this work, our focus is a constrained motion planner for safe object manipulation considering the soft-finger constraints. The goal of the constraints is to eliminate the in-hand sliding of the relatively heavy, long objects, and avoid the end-effector poses where the gravity disturbance goes beyond the limit that the soft finger torque can withstand. 

\section{Overview of the System and Method}
Fig.\ref{workflow} shows an overview of the proposed system. The system comprises two components. The frame box with a red background is the robot manipulation planning component. The frame box with a green background is the human comfortableness estimation component.

The whole system starts with the human comfortableness estimation component. Given the target position of an assembly task, the estimation component first finds the goal poses for each workpiece considering the target position. Then, it samples and searches the workpiece pose space to find a satisfying robot-to-human handover pose at the frame box labeled with marker \textcircled{\small{1}}. With one sampled workpiece pose, the estimation component evaluates the human grasp space embedded in the local coordinate system of the sample. It discretizes the possible human grasp poses in the grasp space and computes the manipulability of the human arm at the discretized grasp poses. The frame boxes labeled with marker \textcircled{\small{2}} and \textcircled{\small{3}} indicates the discretization in the embedded human grasp space and the evaluation of the manipulability in the human arm manipulation space respectively. The grasp pose with the largest manipulability is considered to be most comfortable for the human. The workpiece pose that has the most comfortable grasp pose will be used as the handover pose for the robot. It will be sent to the robot manipulation planning component as the planning goal. The blue arrow in the figure shows the data flow that sends the workpiece pose (planning pose) to the planning component. The planning pose (see the blue frame at the intersection of the two components) bridges the robot planning and the human comfortableness.

On the other hand, the robot manipulation planning component accepts the goal pose of a workpiece and the grasp poses of a human that receives the handover at the frame box labeled with marker \textcircled{\small{4}}. It will use the goal pose of the workpiece as the goal of a multi-modal manipulation planner to find a manipulation motion sequence to move the workpiece from its initial pose to the goal. During the planning, the robot will build a manipulation graph and automatically decide if tools, handover, or other transition states between multiple modalities are necessary. The graph is built in the robot grasp space shown in the frame box labeled with marker \textcircled{\small{5}}. The planning is done by searching the graph and performing motion planning while considering various constraints in the manipulation space labeled with marker \textcircled{\small{6}}. The output of the robot manipulation planning component is a robot motion that moves one workpiece to the handover pose. The motion will be executed by the robot at \textcircled{\small{7}}. After that, the planning component will restart and plan for the next workpiece at \textcircled{\small{8}}, until all workpieces have been handed over to the human.
\begin{figure*}[!htbp]
    \centering
    \includegraphics[width = \textwidth]{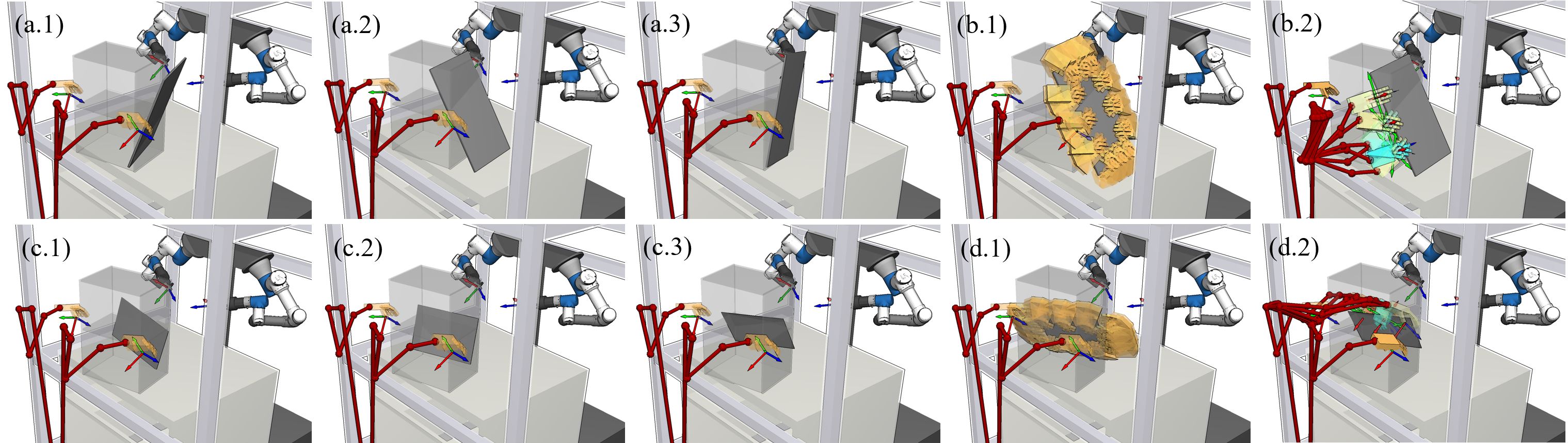}
    \caption{(a.1-3) Three sampled poses for a lateral board of a cabinet. (b.1) The planned grasps for the board pose in (a.2). (b.2) The comfortableness of the available grasps. The warmer color indicates more comfortable grasps for the human right arm. Similarly: (c.1-3) Three sampled pose for a central board of a cabinet. (d.1) The planned grasps for the board pose in (c.3). (d.2) The available grasps ranked according to their comfortableness for the human left arm.}
    \label{determinegoals}
\end{figure*}
The system distributes the subtasks of an assembly to robots and humans by exploiting their advantages and avoiding their disadvantages: although robotic planning and control have been developed for several decades, it is still difficult to flexibly deal with the fine operations needed in assembly tasks using a fully autonomous and general-purpose robot. For this reason, the robot in the system will work on the pick-and-place tasks and provide workpieces to humans. It considers human comfortableness and the assembly pose of the workpieces during the optimization and manipulation planning. On the other hand, the human collaborator will work on fine operations like aligning, fixing, screwing, etc. He or she receives the workpieces from the robot and performs the fine operations without large adjustment of the workpiece poses. In this way, the human can avoid annoying rotations and concentrate on the fine details.

\section{Determining the Planning Goals}
We formulate the comfortableness of a human receiving workpieces considering the manipulability of a human hand as well as the easiness for assembly. The goal is to allow the human to receive the workpieces with a comfortable pose and move the received pieces to assembly positions easily without changing its pose. 

The kinematics of the upper human body and the working range of each arm are modeled based on the average size of a Japanese worker. Given the kinematic parameters, we sample the goal of a workpiece in the collaborative workspace while considering the goal position for the assembly task. The assumption during sampling must meet:
\begin{eqnarray}
\left\{
\begin{aligned}
    ~&||\ln{(\textbf{R}_{sample}\textbf{R}_{goal})}|| < 45^\circ\\
    ~&||\textbf{p}_{sample}-\textbf{p}_{goal}|| < 500~\textit{mm}\\
    ~&\textbf{\textit{M}}_{finished} \cap \textbf{\textit{M}}_{current} = \emptyset,
\end{aligned}\right.
\label{eqcond}
\end{eqnarray}
where $\textbf{p}$ and $\textbf{R}$ represent the position vector and rotation matrix. $\textbf{\textit{M}}$ indicates the mesh model (with transformation) of an object. Fig.\ref{determinegoals}(a.1-3) exemplifies three sampled poses for a lateral board used to assemble a cabinet. They are obtained by random pose sampling following equation \eqref{eqcond}.

Then, for each sampled pose, we plan the candidate grasps to hold the workpiece by treating the human hand as a 1-DoF gripper. The grasp planning algorithm we use was previously published in \cite{wan2016developing}. The algorithm can find the possible grasps performed by a human. During the planning, we estimate the stability of a human grasp by considering soft finger contact. The torque born by the fingers must satisfy the following constraint to avoid slipping.
\begin{equation}
    f_t^2+\frac{\tau_n^2}{e_n^2}\leqslant\mu^2\cdot P^2.
\end{equation}
Here, $f_t$ and $\tau_n$ are the magnitude of the frictional force and moment at a finger contact respectively. $P$ is the pressure force applied to the finger pad along the reverse direction of the contact normal. $\mu$ is the static friction coefficient at the finger contact, it is chosen as 0.8 to approximate human skin \cite{savescu2008technique}. $e_n$ is an eccentricity that relates $f_t$ and $\tau_n$. Under the Winkler elastic foundation \cite{johnson1987contact} with piercing depth $h$, elastic modulus $K$, and relative radii $R^{'}$ and $R^{''}$ for the finger and the object in contact, $e_n$ equals
\begin{equation}
    e_n=\frac{8}{15}\sqrt{a\cdot b}=\frac{16}{15}\sqrt{\frac{P\cdot h\cdot(R^{'}\cdot R^{''})^\frac{1}{2}}{k\cdot\pi}}.
\end{equation}

For each of the planned grasps, we solve the IK (Inverse Kinematics) using the human kinematic parameters to estimate if a human may hold the workpiece using the grasp. Then, for an IK-feasible grasp, we estimate its comfortableness by using the inverse of the condition number of a human arm's Jacobian matrix \cite{kim1991dexterity}\cite{salisbury1982articulated}:
\begin{equation}
    Q_{g_i}=\frac{\sigma_{min}}{\sigma_{max}},
    \label{humanmanip}
\end{equation}
where $\sigma_{min}$ and $\sigma_{max}$ are the smallest and largest singular values of the human arm's Jacobian. The subscript $_{g_i}$ indicates a grasp configuration. The inverse of the condition number of J, instead of the classical manipulability defined by Yoshikawa \cite{yoshikawa1985manipulability} is used to estimate the comfortableness since this value is non-dimensional and independent of the scale of an arm.

Fig.\ref{determinegoals}(b.1) and (b.2) exemplify the planned grasps for the workpiece pose shown in Fig.\ref{determinegoals}(a.1) and marked their availability and comfortableness respectively. The hands with a warmer color in Fig.\ref{determinegoals}(b.2) indicate the grasp is more comfortable.

Following the sampling and evaluation process, we can get a list of candidate goal poses ordered using the most comfortable of the respective candidate grasps. The list will be sent to the robot for motion planning. The robot will iterate through the candidate goal poses in the list until it finds a feasible motion to pose the workpiece to a goal.

\section{Planning the Manipulation Motion}
\label{planmanipmotion}
\begin{figure}[!tbp]
	\centering
	\includegraphics[width=\columnwidth]{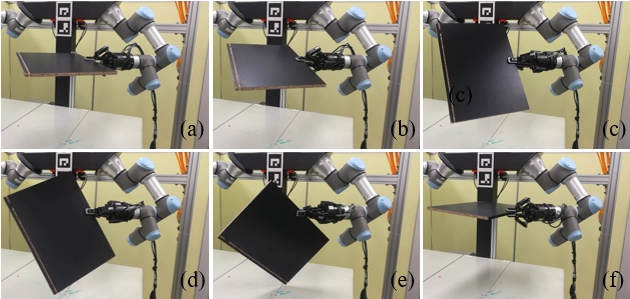}
    \caption{Motion sequence of a robot trying to flip a long and heavy board. The flipping motion is planned without considering any constraints. The shortest path for this task is to rotate $180^{\circ}$ about the approach vector of the end effector. Such a motion results into a rotational slip and    the in-hand pose is changed ((a)(b)(c) $vs.$ (f)(e)(d)).}
	\label{slipfailure}
\end{figure}

In this section, we explain the details of the robot motion planning under the soft finger stability. The workpieces, namely the cabinet boards studied in this paper, are long and heavy. Manipulating them is challenging because the board may experience undesirable slipping when the robot arm passes through certain poses. Fig.\ref{slipfailure} illustrates a situation in which a board slipped due to a bad plan. In this case, no constraints on motion planning are considered. The in-hand pose is changed and the execution fails as the robot motion increases the gravity torque born by the grasped board. To avoid such failure, we implement a motion planner for safe object manipulation and consider the soft-finger constraints. We especially incorporate a constraint relaxation strategy for efficient motion planning. The details are as follows.

\subsection{Modeling the slip motion}
Slip motion occurs under the influence of two main factors: (i) The gravity torque on the object. (ii) The friction torque between the object and the robotic fingertips. As described by Eq.\eqref{eq5}, the gravity and friction torques play opposite roles in the way how the object slips in the robot hand. If the gravity torque $T_g$ exceeds the friction torque $2T_f$ exerted by both of the fingertips, the object will start to slip with an angular acceleration:
\begin{equation}
    \ddot{\phi} = T_g/\mathcal{I} - 2T_f/\mathcal{I}
\label{eq5}
\end{equation} 
where
\begin{gather}
    T_g = \dfrac{mg}{2} sin(\theta) sin(\phi) (Obj_{CoM_{rel-EE}}  \hspace{1mm} sin(\phi)) \nonumber \\
    + \dfrac{mg}{2} sin(\theta) cos(\phi) (EE_{length}+Obj_{CoM_{rel-EE}} \hspace{1mm} cos(\phi)), 
\raisetag{2.5\normalbaselineskip}
\label{eq6}
\end{gather}
Fig.\ref{slipmodel} shows the definition of the symbols used to compute the gravity torque in Eq.\eqref{eq6}. The end-effector inclination angle $\theta$ is defined concerning the $\textit{\textbf{z}}$ axis of the gripper's local reference frame. As is shown in the figure, the angle $\theta$ is essentially the rotation of the end-effector around the $\textit{\textbf{z}}$ axis. When this inclination angle $\theta$ is larger than a threshold $\theta_{c}$, the object starts to slip. It may finally rest at a pose with a relative angle $\phi_{slip}$ to the reference frame. This $\phi$ angle is defined as the in-hand slip angle. Its value is:
\begin{figure}[!tbp]
	\centering
	\includegraphics[width=0.65\columnwidth]{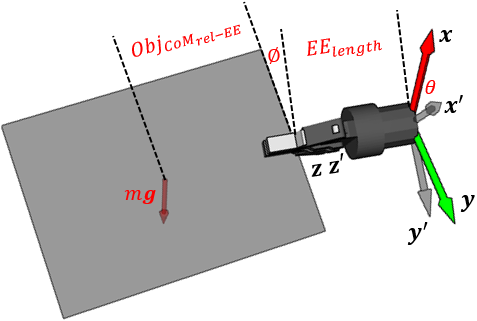}
    \caption{A grasped board after experiencing in-hand rotational slip of value $\phi$ due to end-effector inclination $\theta$. The red vector in the middle of the board represents the gravity vector at the object center of mass.}
	\label{slipmodel}
\end{figure}
\begin{eqnarray}
\phi = 
\left\{
\begin{aligned}
    ~& 0 \hspace{1.0cm} \theta  <\theta_{c} \\
    ~& \phi_{slip} \hspace{0.5cm} \theta >\theta_{c}
\end{aligned}\right.
\label{eq7}
\end{eqnarray}
\begin{figure*}[!htbp]
    \centering
    \includegraphics[width = \textwidth]{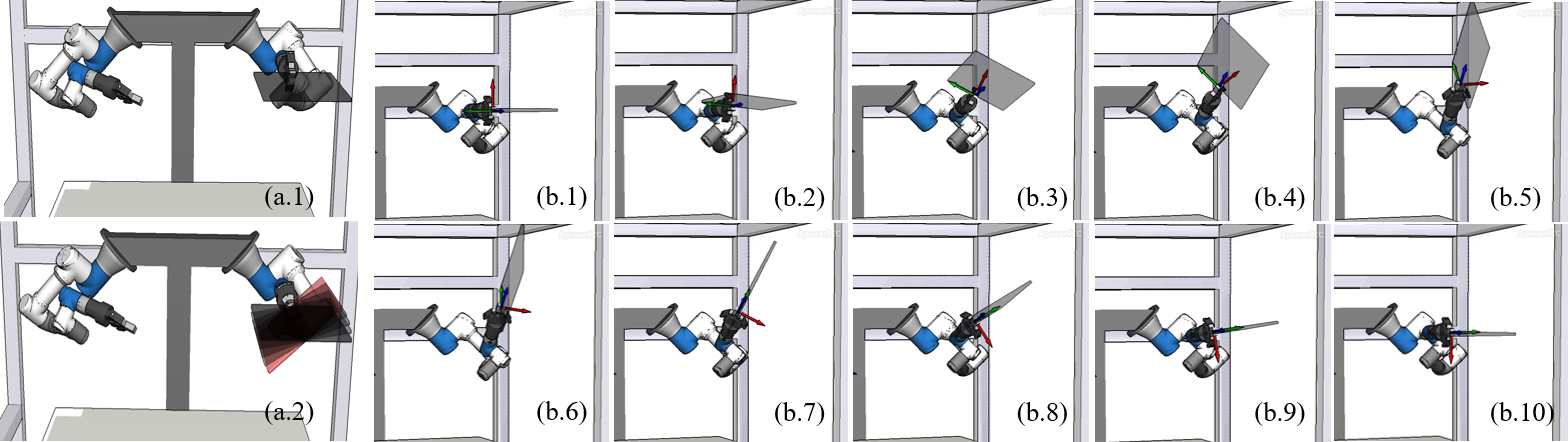}
    \caption{(a.1) The object pose and the grasp. (a.2) The constraint relaxation procedure, the object is inclined in steps between $0^{\circ}$ and $90^{\circ}$ degrees and the gravity torque at each inclination is compared to the grasp friction torque. If the gravity torque becomes higher, the constraint is set. The critical inclination of the grasped object is highlighted in red. (b1-10) The left-arm motion that performs the flipping task.}
    \label{ctrelaxation}
\end{figure*}
Eliminating the in-hand slip angle is the objective of our study. The end-effector inclination makes the object weight $mg$ unstable in the end-effector grasp. It increases the gravity torque born by the object by an amount equivalent to $sin(\theta)$.

The $EE_{length}$ notation in Eq.\ref{eq6} is the length of the end-effector. $Obj_{CoM}$ is the distance from the object's center of mass to the grasp point. The gravity torque can be calculated at any end-effector inclination and object slip angle following the equation.

Eq.\eqref{eq6}-\eqref{eq7} show that the variable with the most significant effect on the gravity torque is the inclination angle $\theta$. Thus the variable is taken into account as the constraint variable in our motion planner to avoid the rotational slip.

\subsection{Gravity torque based constraint}
Following the model of the slip, we add a gravity torque based constraint to the motion planner by controlling the inclination of the end-effector. The constraint is formulated as an angle inequality condition in the motion planning algorithm. The angle between the opening direction of the end-effector and the gravity vector is used to measure the inclination. Here, we exemplify two extreme cases of the end-effector's inclination using this angle to explain its utility. The safest case against slip is when the angle between the opening direction and the gravity is $0^{\circ}$. Such a case is shown in Fig.\ref{slipfailure}(a). No slip occurs as the whole weight of the object is supported by the end-effector. The worst case is when the angle is $90^{\circ}$. In this case, the opening direction is normal to the gravity, as shown in Fig.\ref{slipfailure}(e). The whole weight of the object is contributing to the gravity torque, and the object is very likely to slip in-hand.

\subsection{Constraints relaxation criteria}
\label{ctrelaxation}
Constraint relaxation is applied to avoid adding overly strict constraints and results in narrow passages in the planning space. As explained in the previous subsection, the grasped object is mostly secured against slipping when the angle between the opening direction of the end-effector and the gravity is $0^{\circ}$. Considering this extreme, the constraint relaxation searches around $0^{\circ}$ to find a boundary $0^{\circ}\pm \theta$ in which the object is still strongly secured against slip. The constraint checking component in our system examines for this range and sets up the relaxation limit in a pre-processing step. Fig.\ref{ctrelaxation} shows an example of the pre-processing step as well as the resulted motion considering the relaxed constraint. Fig.\ref{ctrelaxation}(a.1) is the safest pose where no end-effector inclination exists. To examine the allowable relaxation limit, the inclination angle is increased gradually in evenly discretized steps. The gravity torque increases along with the inclination according to Eq.\eqref{eq6}. At each step, the calculated gravity torque is compared with the maximum friction torque that the end-effector's soft finger pad can provide. The constraint relaxation limits $\pm \theta$ is set to be an inclination angle at which the gravity torque exceeds the maximum friction torque. This process is illustrated in Fig.\ref{ctrelaxation}(a.2) with the resulting relaxation limit angle highlighted in dark red color. Fig.\ref{ctrelaxation}b(1-10) is the motion sequence that successfully flips the same board in Fig.\ref{slipfailure} considering the relaxed constraint. During the motion planning, each randomly sampled robot pose is required to satisfy the constraint. If the constraint is violated, the sample is discarded and a new search for constraint-satisfying poses is continued until the goal is reached.

\section{Handover}
\subsection{Handover between the robotic arms}
There are two handovers in the planner. One is the handover between the robotic arms. The other is the handover between the robot and the human. In this subsection, we discuss the handover between the robotic arms. Since the boards are thin and are not directly graspable by the robotic grippers, a suction cup tool \cite{chen2019humanoids} is employed for easy pick-up. Following the algorithms presented in \cite{chen2019humanoids}, one of the robot arms will manipulate the suction tool, while the other is used to manipulate the boards. The suction tool is used to pick-and-place boards. Each board is picked up by the arm holding the suction-cup tool, and moved from its initial pose on the robot work table to an intermediate pose, namely the robot-robot handover pose. The other arm then receives the board from the handover pose and manipulates it. While the pick-up position is always at the center of mass of each board, the rotation of the tool, the handover pose, as well as the receiving grasping configuration are optimized together with the handover between the robot and the human. The details are discussed in the next subsection.

\begin{figure*}[!htbp]
    \centering
    \includegraphics[width = \textwidth]{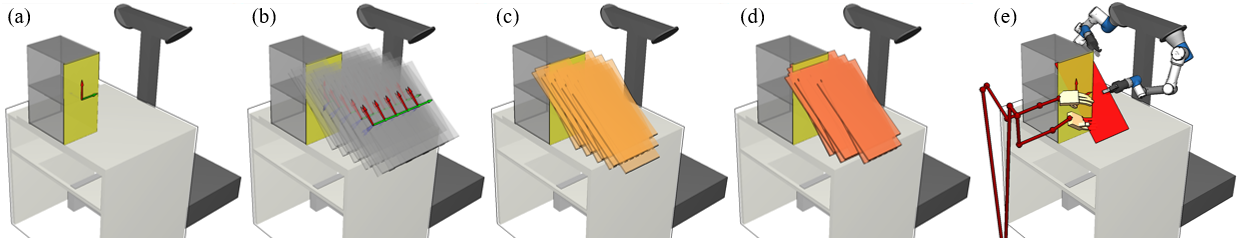}
    \caption{The process of the robot-human handover pose determination. (a) A cabinet board at its target assembly pose. (b) With reference to the assembly pose, the initial candidate poses for the handover are sampled by moving the object away from its assembly pose and rotating randomly. (c) The candidate poses are checked for their comfortableness to the human, and the satisfying poses are shown. (d) The remaining poses are further checked to find shared grasps for robot-robot handover. The poses with shared grasps are kept and shown. (d) The candidate poses that satisfy both of the previous conditions are sorted according to their distance from the assembly target pose, and the nearest candidate is selected to be the robot-human handover pose. The figure also illustrates the best human receiving gesture at this board pose.}
    \label{goaldetermine}
\end{figure*}
\begin{figure}[!htbp]
	\centering
	\includegraphics[width=\columnwidth]{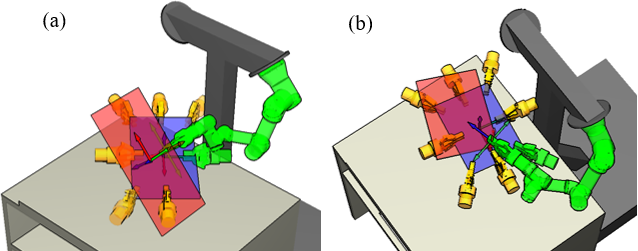}
    \caption{Finding the shared grasps for robot-robot handover and the robot-human handover. The robot-robot handover is shown in blue. The robot-human handover is shown in red. The unavailable grasps at the robot-robot handover are highlighted in orange. The available ones are highlighted in green. The shared grasps between the two handovers are also illustrated using green color, with the whole arm additionally drawn.}
	\label{startgoalshared}
\end{figure}

\subsection{Handover to the human}
The handover between the robot and a human is optimized by considering the comfortableness of the human's receiving posture. The optimization process is shown in Fig.\ref{goaldetermine}. First, the candidate poses for robot-human handover are sampled with a bias towards the target assembly pose of the board. The bias involves moving the board away from its target assembly pose and rotating it randomly. The process is illustrated in Fig.\ref{goaldetermine}(a, b). The next procedure is to estimate the comfortableness of the sampled poses using comfortableness quality for the human (recall Eq.\eqref{humanmanip}). The available human grasps for each sampled pose are evaluated and ranked by using the quality index. The satisfying handover poses, namely the handover poses whose largest comfortableness value is larger than a threshold, are kept for further use. Eq.\eqref{eqlist} shows the process. Here, $\binom{\textbf{p}}{\textbf{R}}^{rh}_i$ indicates a candidate handover pose. The superscript $^{rh}$ indicates a robot-human handover. The subscript $_{g^{rh}_{i_j}}$ indicates the available human grasps associated with the $i$th candidate handover pose, namely $g^{rh}_{i_j}\in G(\binom{\textbf{p}}{\textbf{R}}^{rh}_i)$. The equation requires the $max()$ value of all candidate grasps to be larger than a threshold. An example of the result is shown in Fig.\ref{goaldetermine}(c), where the light orange boards illustrate the kept poses.
\begin{eqnarray}
    S=\{\binom{\textbf{p}}{\textbf{R}}^{rh}_i|max(Q_{g^{rh}_{i_j}})>threshold,\nonumber \\
    g^{rh}_{i_j}\in G(\binom{\textbf{p}}{\textbf{R}}^{rh}_i)\}
    \label{eqlist}
\end{eqnarray}

Then, the set of the human comfortable poses are checked for shared grasps with the start pose of the board at which the robot-robot handover is carried out. Eq.\eqref{eqlist2} shows the details. Here, the superscript $^{rr}$ indicates a robot-robot handover. The obtained set $\mathbb{S}$ of poses satisfy both the human comfortableness and the robot reachability. 
\begin{eqnarray}
    \mathbb{S}=\{\binom{\textbf{p}}{\textbf{R}}^{rh}_i|g^{rr}_i\in G(\binom{\textbf{p}}{\textbf{R}}^{rh}_i), g^{rr}_i\in G(\binom{\textbf{p}}{\textbf{R}}^{rr}_i), \nonumber \\\binom{\textbf{p}}{\textbf{R}}^{rh}_i\in S\}
    \label{eqlist2}
\end{eqnarray}
These poses are shown in Fig.\ref{goaldetermine}(d). Fig.\ref{startgoalshared} shows two examples of the shared robot grasps between the robot-robot handover pose and the robot-human handover pose. The robot-robot handover pose of the board is shown in blue. The robot-human handover pose is shown in red. The available grasps at the robot-robot handover pose are illustrated in green. The unavailable ones are shown in orange. The resulting shared grasp, by which the board will be passed and received, is denoted by the green arm at the goal robot-human handover.

In the last step, the candidate poses are sorted according to a quality metric $Q_{\mathbb{S}}$ which measures their distances from the target assembly pose. The nearest pose is selected as the goal robot-human handover pose according to relationship Eq.\eqref{eq8}. The selected goal robot-human handover pose is illustrated in Fig.\ref{goaldetermine}(e). The best human grasps at this pose and the robot pose that hands the board to the human are also illustrated for better understanding. Note that the nearest pose is not necessarily reachable by the robot. During execution, if the path between the start pose and the nearest pose is blocked because of the planning constraints, e.g. collision detection, the next nearest pose is considered and so on. 
\begin{equation}
    \{\min{(Q_{\mathbb{S}})} \mid Q_{\binom{\textbf{p}}{\textbf{R}}^{rh}_i} = \binom{\textbf{p}}{\textbf{R}}^{rh}_{target}-\binom{\textbf{p}}{\textbf{R}}^{rh}_{i}, \binom{\textbf{p}}{\textbf{R}}^{rh}_{i} \in \mathbb{S} \}
\label{eq8}
\end{equation}

Once the robot arm brings the cabinet board to the desired robot-human handover pose, the robot starts waiting for human reception. Force/torque sensors installed at the wrists of the robot arms are used to check for the reception action. When the changes of forces in the passing arm are detected, the robot opens the gripper to release the board to the human collaborator. 
\begin{figure}[!t]
	\centering
	\includegraphics[width=0.9\columnwidth]{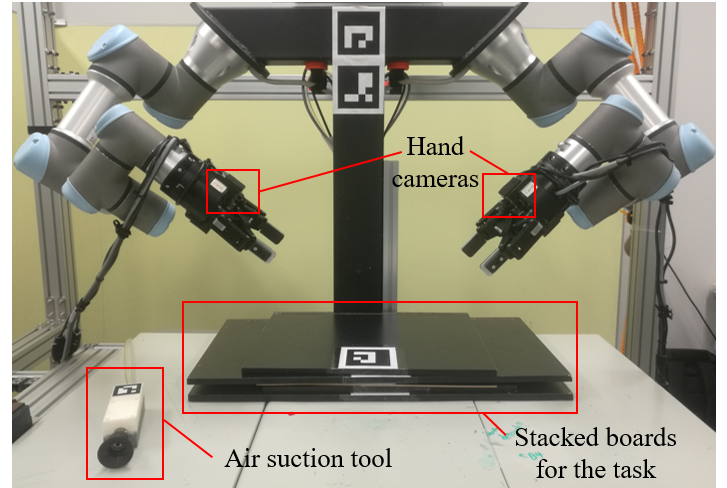}
    \caption{The experimental systems and setups. A dual UR3 robot is used to perform the collaborative assembly with the human. At the end of each arm, there are two Robotiq FT300 6DoF F/T sensors and two Robotiq 2F-85 grippers. Two hand cameras are installed at the two sides of the grippers to detect the boards and the suction cup used for pick-and-place tasks.}
	\label{expsetup}
\end{figure}
\begin{table}[!t]
\setlength{\tabcolsep}{1.2em}
\begin{center}
\begin{tabular}{@{}cccccc@{}}
\toprule\toprule
\multirow{2}{*}{Board} & \multirow{2}{*}{No.} & \multicolumn{3}{c}{Dimensions (mm)}                                                    & \multirow{2}{*}{Mass (kg)} \\ \cmidrule(lr){3-5}
                       &                      & \multicolumn{1}{l}{Length} & \multicolumn{1}{l}{Width} & \multicolumn{1}{l}{Thickness} &                            \\ \midrule 
Large                  & 2                    & 390                        & 288                       & 10                            & 0.8                        \\ \midrule
Medium                 & 3                    & 587                        & 295                       & 10                            & 1.8                        \\ \midrule
Small                  & 2                    & 397                        & 280                       & 3                             & 0.22                       \\ \bottomrule\bottomrule
\end{tabular}
\caption{The different types of cabinet boards, the number of boards for each type, their dimensions, and masses.}
\label{table:1}
\end{center}
\end{table}
\begin{figure*}[!t]
    \centering
    \includegraphics[width = \textwidth]{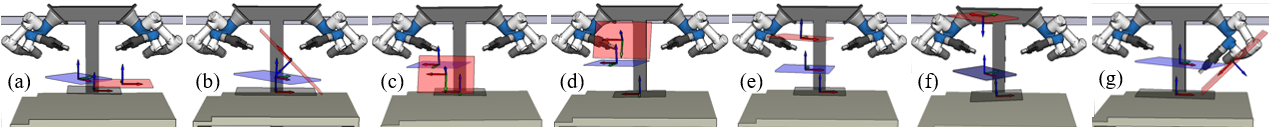}
    \caption{The optimized robot-robot handover poses and robot-human handover poses for the seven boards. The initially detected board poses are illustrated in grey. The robot-robot handover pose is illustrated in blue.    The optimized robot-human handover pose is shown in red.}
    \label{exp_initstartgoal}
\end{figure*}
\begin{figure*}[!t]
    \centering
    \includegraphics[width = \textwidth]{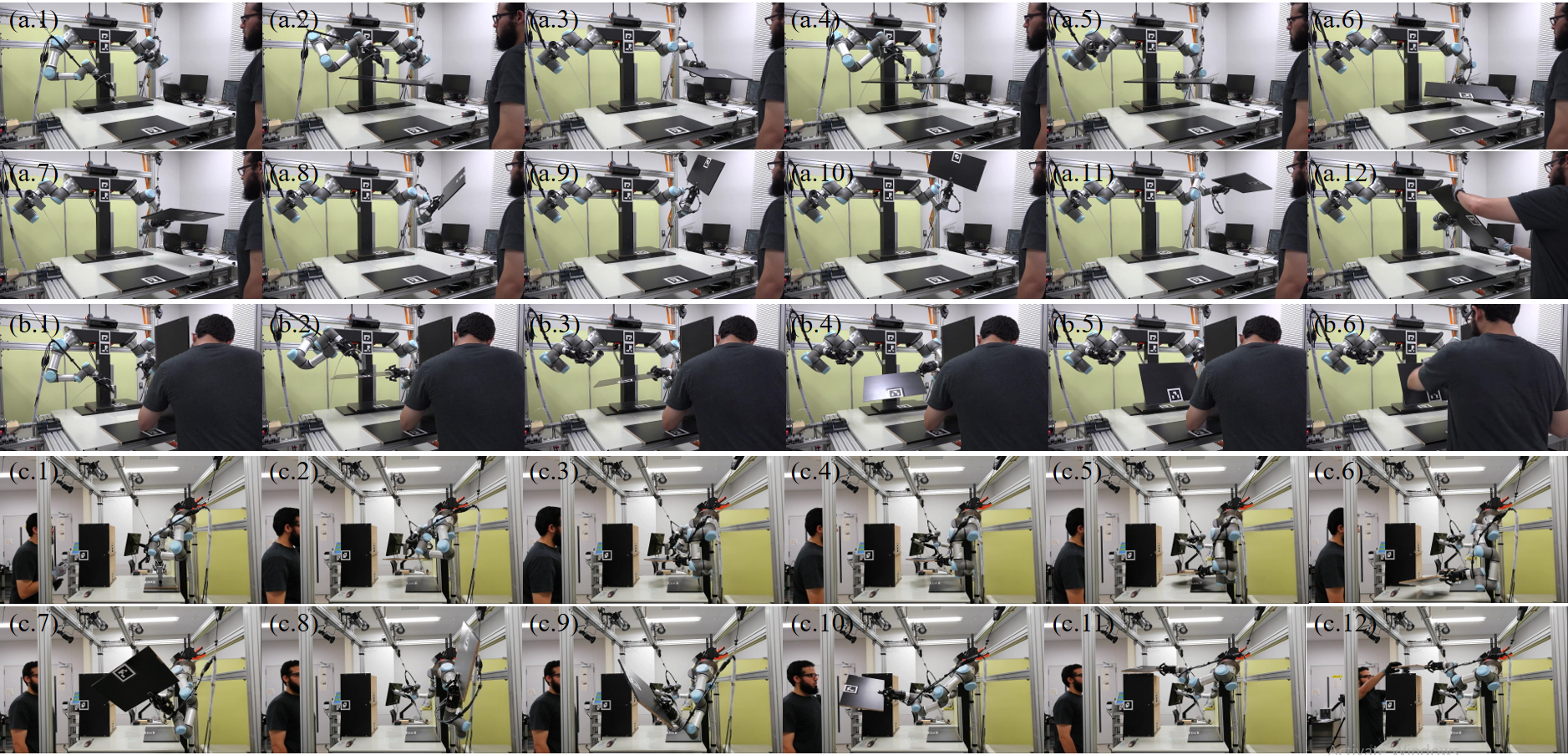}
    \caption{(a) The motion sequence to manipulate the second board. This is the right lateral board of the cabinet. It has the largest dimensions and thus a tight slip constraint. The arm poses its arm up to minimize the gravity torque on the board while making the rotation required to flip the board to the optimized robot-human handover pose. (b) The motion sequence of the third board. In this situation, no flipping is required. The figure sequence focuses on the reception of the human. (c) The motion sequence of the sixth board. This is the uppermost board of the cabinet. At this point, a large portion of the cabinet has been finished, making the workspace narrow. The sequence highlights the planner's capability in fulfilling the constraints while the workspace is crowded by the finished part.}
    \label{exp_sequences}
\end{figure*}

\section{Experiments and Analysis}
We perform a real-world collaborative human-robot assembly task of a cabinet as well as multiple simulations to verify the utility of our developed system. The experimental systems and setups are shown in Fig.\ref{expsetup}. The dual-arm robot is made of two UR3 arms produced by Universal Robots Ltd. At the end of each arm, there are two Robotiq FT300 6DoF F/T sensors and two Robotiq 2F-85 grippers. Hand cameras are installed at both sides of the grippers for visual detection. The target object is a cabinet constituting seven boards of three different types. The dimensions of them are shown in Table \ref{table:1}. In the beginning, the cabinet boards are stacked in front of the robot. A suction tool is randomly placed beside the boards for easy pick-up. 
The results of the experiments are summarized in a supplementary video attached to this paper. Readers may also find the video online at https://youtu.be/t\_-89-N\_RgM. Along with the seven boards, there are seven times of pick-up, robot-robot handover, and robot-human handover during the collaborative assembly. The optimized robot-robot handover poses and robot-human handover poses for all seven times are illustrated in Fig.\ref{exp_initstartgoal}. 

For the first board, the robot picks the suction tool with the right arm, uses it to pick the cabinet baseboard (medium size) from its initial pose,  and hands it over to the left arm. The board is manipulated by the left arm and posed to the optimized robot-human handover pose. The constraint definition component applies a relaxation angle of $\theta = 62^{\circ}$ to the constraint of the motion planner when this board is grasped along its transverse axis. This constraint provides more freedom in the planning as the angle between the opening direction of the end-effector and the gravity vector is permitted in the range of $0^{\circ} \pm 62^{\circ}$. If the same board is grasped along its longitudinal axis (recall Fig.\ref{ctrelaxation}(b.1-2)), the applied relaxation becomes $\theta = 40^{\circ}$ (which is reasonable as the center of mass of the object becomes farther from the contact point), allowing the gravity torque to build faster at smaller inclination angles. The first sub-task is relatively simple because there is no requirement for flipping the board to fit for the optimized robot-human handover pose.

For the second board, the robot follows a similar procedure to hand over the cabinet lateral board (large size) to the human. This board is more difficult (the constraint should be more strict) since it has a larger dimension and is much heavier than the first one, which makes the gravity torque effect and slipping significant even at a small inclination angle. Also, the board is required to be flipped to reach the optimized goal pose for the robot-human handover. As expected, the constrained checking component suggests a more tight constraint to secure the object against in-hand slip. The calculated safe range for the inclination angle for this board is $0^{\circ} \pm 22^{\circ}$. The system successfully finds a motion for the robot to manipulate the board without slip under this constraint. The sequence of the robot motion is shown in Fig.\ref{exp_sequences}(a.1-12) for the reader's convenience.

The third board has a much smaller weight compared to the previous ones. The constraint checking component determines that there is no need to constrain the arm poses while manipulating this board. The permitted range of inclination is, therefore, the whole range of $0^{\circ} \pm 90^{\circ}$. This constraint allows the robot to be at an arbitrary pose during manipulation. The motion sequence about board is shown in Fig.\ref{exp_sequences}(b.1-6).

The fourth board is the uppermost board of the cabinet. It is a medium-size one like the first board. At this stage of the assembly task, a large portion of the cabinet has been assembled and the workspace becomes narrower because of the finished parts. Also, although the constraint defined for this case is the same as the first experiment $\theta = 62^{\circ}$, the board is required to be flipped to fit for its optimized robot-human handover pose. For these reasons, this board is considerably difficult to manipulate. The robot needs to perform the flipping task while tracing the collision-free workspace. Fig.\ref{exp_sequences}(c.1-12) shows the motion sequence for this board.

The remaining three boards are similar to the previous ones. Thus they are not discussed in detail. Please see the attached video for details. 

To further understand how the collaborative system improves the assembly task, we add more experiments to compare the performance of the assembly with either a human or a robot on its own. The human is more flexible in dealing with task variations and can carry out fine manipulation tasks which are less physically demanding. However, handling flipping and reorienting tasks can be tiresome. On the other hand, a robot can have good performance in pick-and-place and repetitive tasks. However, it has difficulty in handling fine manipulation tasks. Specialized tools, as well as complicated control and sensing, are required for fine operations. Also, a robot is highly constrained by its kinematic structure. It cannot reach beyond its workspace and may encounter singularities inside the workspace. All these factors limit a robot's ability to manipulate objects. Meanwhile, both human and robots will lose time by switching tasks. They suffer from the need for more hands to get the assembly task done.  

Table \ref{table:2} presents a comparison of the required human physical workload for the assembly task using our developed approach and a human-alone approach. The comparison is based on the Ergonomic Assessment Worksheet (EAWS) \cite{schaub2013european} while considering the activities described in \cite{malaise2019activity}. The results show that our approach eliminates the need of bending forward as the board is presented to the human in a pose that considers his position in front of the robot. Meanwhile, the shoulder work is decreased as the robot hands over the board in an already flipped pose. It can be used directly for assembly. The human exerts only a light shoulder work for placing the board and assembling it. Reaching and picking actions are both reduced and are made more comfortable for the human. The idle human activity state is increased as the robot shares some workload. The fine manipulation by the human for screwing remains the same.
\begin{table}[!tbp]
\setlength{\tabcolsep}{.25em}
\centering
\begin{tabular}{@{}lll@{}}
\toprule
\toprule
Activity        & \multicolumn{1}{c}{Human-robot} & \multicolumn{1}{c}{Activity description}                                                                                                                        \\ \midrule
\multicolumn{3}{c}{\dunderline{.57pt}{Main human posture}}                                                                                                                                              \\ [2mm]
Standing          & The same    & Ends when human start to move                                                                                                                            \\ \midrule
\multicolumn{3}{c}{\dunderline{.57pt}{Torso and arms configuration}}                                                                                                                                     \\[2mm]
\begin{tabular}[l]{@{}l@{}}Bent\\forward\end{tabular} & Eliminated  & Torso bending angle: $20^{\circ}\sim60^{\circ}$    \\ [2mm]
\begin{tabular}[l]{@{}l@{}}Shoulder\\work\end{tabular} & Decreased   & \begin{tabular}[l]{@{}l@{}}Elbow at or above shoulder level and \\ hands at or below head level\end{tabular}                                            \\ \midrule
\multicolumn{3}{c}{\dunderline{.57pt}{Goal-oriented action}}                                                                                                                                         \\ [2mm]
Reaching          & \begin{tabular}[l]{@{}l@{}}More\\ comfortable\end{tabular}   & \begin{tabular}[l]{@{}l@{}}Moving an arm towards a target with\\ no object in hand\end{tabular}                                                         \\[3mm] \renewcommand{\arraystretch}{5}
Picking           & \begin{tabular}[l]{@{}l@{}}More\\ comfortable\end{tabular}   & \begin{tabular}[l]{@{}l@{}}Picking up an object, starts when \\touching the object, ends when arm \\stops motion with respect to the body\end{tabular} \\ [5mm]
Placing           & The same    & \begin{tabular}[l]{@{}l@{}}Moving an arm towards a target, with \\ object in hand\end{tabular}                                                          \\ [2mm]
\begin{tabular}[l]{@{}l@{}}Fine \\ manipulation\end{tabular}  & The same    & Dexterous manipulation of an object                                                                                                                     \\ [2mm]
Idle              & Increased   & Not doing anything with hands                                                                                                                           \\ \bottomrule\bottomrule
\end{tabular}
\caption{Comparison of the required ergonomic activities for the assembly task -- The human-robot collaboration approach \textit{vs.} The conventional human-alone approach.}
\label{table:2}
\end{table}

The real-world experiments and the comparison show that the proposed system is effective, efficient, and can help humans to finish assembly task comfortably. 


\section{Conclusions and Future Work} 
In this paper, we propose a human-in-the-loop robotic manipulation planner for collaborative assembly. The system distributes the subtasks of an assembly to robots and humans by exploiting their advantages and avoiding their disadvantages. It incorporates multiple components to improve overall efficiency. From the human perspective, the tasks are becoming less labor-intensive and more comfortable. From the robot side, the manipulated objects become safer against undesirable slip with constrained motion planning and relaxation. The experiments and analysis confirmed these claims. The next work will be updating the human model online using vision sensors to adapt to a wide range of users.

\bibliographystyle{ieeetr}
\bibliography{reference}

\end{document}